\pdfoutput=1

\documentclass[11pt]{article}

\usepackage[final]{EMNLP2023}

\usepackage{times}
\usepackage{latexsym}
\usepackage{mdframed}
\usepackage[T1]{fontenc}

\usepackage[utf8]{inputenc}

\usepackage{microtype}

\usepackage{inconsolata}

\usepackage{caption}
\usepackage{subcaption}
\usepackage{enumitem}
\usepackage{flushend}
\usepackage{balance}
\usepackage{lineno}
\usepackage{url}
\usepackage{amsmath,amssymb,amsfonts}
\usepackage{algorithm}
\usepackage{algorithmic}
\usepackage{graphicx}
\usepackage{textcomp}
\usepackage{xcolor}
\usepackage{hyperref}
\usepackage{colortbl}
\usepackage{amsthm}
\usepackage{float}
\usepackage{multirow}
\usepackage{multicol}
\usepackage{color}
\usepackage{bm}
\usepackage{bbm}
\newtheorem{theorem}{Theorem}[section]

\newtheorem{prop}[theorem]{Proposition}

\newtheorem{remark}{Remark}

\usepackage{pifont}

\newcommand{\bx}{\mathbf{x}}
\newcommand{\by}{\mathbf{y}}
\newcommand{\bz}{\mathbf{z}}

\usepackage{array}
\newcolumntype{L}[1]{>{\raggedright\let\newline\\\arraybackslash\hspace{0pt}}m{#1}}
\newcolumntype{C}[1]{>{\centering\let\newline  \\\arraybackslash\hspace{0pt}}m{#1}}
\newcolumntype{R}[1]{>{\raggedleft\let\newline \\\arraybackslash\hspace{0pt}}m{#1}}

\newcommand{\mypara}[1]{{\smallskip \noindent \bf #1}\hspace{0.1in}}

\DeclareMathOperator*{\argmax}{argmax}

\usepackage[utf8]{inputenc} 
\usepackage[T1]{fontenc}    
\usepackage{hyperref}       
\usepackage{url}            
\usepackage{booktabs}       
\usepackage{amsfonts}       
\usepackage{nicefrac}       
\usepackage{microtype}      
\usepackage{xcolor}         
\usepackage{comment}
\usepackage{arydshln}

\title{FastGAS: Fast Graph-based Annotation Selection for In-Context Learning}

\author{Zihan Chen, Song Wang, Cong Shen, Jundong Li\\
Department of ECE, University of Virginia, Charlottesville, VA, USA\\
\{brf3rx,sw3wv,cong,jl6qk\}@virginia.edu}

\begin{document}
\maketitle
\begin{abstract}

In-context learning (ICL) empowers large language models (LLMs) to tackle new tasks by using a series of training instances as prompts. Since generating the prompts needs to sample from a vast pool of instances and annotate them (e.g., add labels in classification task), existing methods have proposed to select a subset of unlabeled examples for annotation, thus enhancing the quality of prompts and concurrently mitigating annotation costs. However, these methods often require a long time to select instances due to their complexity, hindering their practical viability. To address this limitation, we propose a graph-based selection method, FastGAS, designed to efficiently identify high-quality instances while minimizing computational overhead.
Initially, we construct a data similarity graph based on instance similarities. Subsequently, employing a graph partitioning algorithm, we partition the graph into pieces. Within each piece (i.e., subgraph), we adopt a greedy approach to pick the most representative nodes. By aggregating nodes from diverse pieces and annotating the corresponding instances, we identify a set of diverse and representative instances for ICL. Compared to prior approaches, our method not only exhibits superior performance on different tasks but also significantly reduces selection time. In addition, we demonstrate the efficacy of our approach in LLMs of larger sizes.

\end{abstract}

\section{Introduction}\label{sec:intro}
Recent advances in natural language processing heavily leverage large language models, exemplified by models such as GPT-3~\cite{brown2020language}. Among them, in-context learning (ICL) emerges as a promising direction in this field. ICL adapts specific tasks with just a few instances as prompts, offering a viable alternative to traditional supervised fine-tuning~\cite{liu2023pre}. 
The performance of ICL is intricately tied to the effectiveness of the prompt surface, encompassing factors such as instance selection and the sequence of demonstration instances~\cite{zhao2021calibrate,dong2022survey,lu2021fantastically, wang2023knowledge}. In this study, we focus on instance selection and explore novel solutions to reduce manual annotation costs while maintaining high in-context learning performance.

Previous research underscores the importance of retrieving prompts from a vast set of annotated instances to achieve optimal performance~\cite{liu2021makes, rubin2021learning}. In particular, performance is shown to improve significantly when choosing in-context examples similar to each test input~\cite{liu2021makes}. However, addressing the unique requirements of different test instances requires a large set of annotated examples, incurring significant human and financial resources.

To mitigate annotation costs, previous efforts have sought to identify a small number of unlabeled instances for annotation~\cite{su2022selective,zhang2023ideal}. The objective is to select diverse and representative instances, where representativeness aids in finding similar demonstrations for different test instances, while diversity broadens the overall coverage. Despite their superiority over random selection, these methods have specific drawbacks. For example, Vote-$k$~\cite{su2022selective} emphasizes diversity but adds inference costs due to predictions on unlabeled data. IDEAL~\cite{zhang2023ideal} employs influence-driven selective annotations, drawing inspiration from influence maximization in social graphs. However, both methods struggle to balance diversity and representativeness, leading to suboptimal performance.

\begin{figure}[htbp]
\centering
\includegraphics[width=\linewidth]{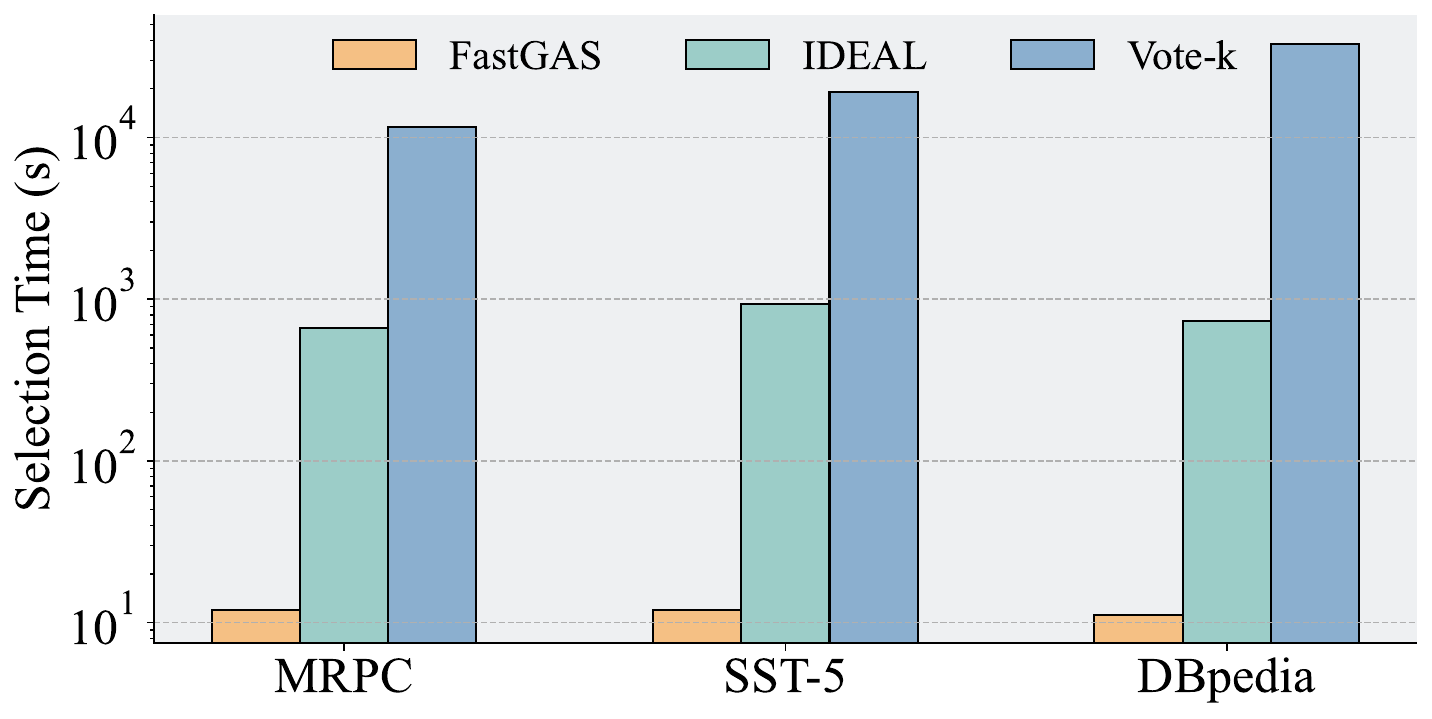}
\caption{Comparison of our method and two baselines on three classification tasks (MPRC, SST-5, and DBpedia) with respect to time consumption during subset selection. The annotation budget is $18$. The y-axis represents the time consumption with a log scale. Notably, our method significantly reduces the time cost in comparison to both baseline methods.}
\vspace{-0.2in}
\label{fig:time_18}
\end{figure}

Furthermore, a notable shortcoming of existing methods is their computational inefficiency. The precise calculation process (e.g., iteratively searching the entire dataset) results in high computational costs, making them less practical for real-world applications. Figure~\ref{fig:time_18} illustrates the time required by our method and two baseline methods, Vote-$k$ and IDEAL, under the same hardware conditions and annotation budgets. It is observed that both baselines necessitate at least $500$ seconds to select a subset for DBpedia and SST-5 tasks. Remarkably, for the DBpedia task, Vote-$k$ exceeds $30,000$ seconds (over eight hours) to select just 18 instances. As the annotation budget grows, the time needed by these methods to perform the selection process can increase exponentially (See Section~\ref{sec:time}), further constraining their applicability in real-world settings. In our pursuit of an efficient and effective selective annotation method, we pose the fundamental question: \textit{Can we identify a set of \textbf{diverse} and \textbf{representative} instances with high \textbf{efficiency}?}

Answering this question, we propose \textbf{FastGAS}, a \underline{\textbf{F}}ast \underline{\textbf{G}}raph-based \underline{\textbf{A}}nnotation \underline{\textbf{S}}election method that works in an unsupervised manner. We first build a data similarity graph based on the similarity among unlabeled data. We then select instances for annotation based on the data similarity graph. In particular, our method focuses on the following three properties of selected instances: 

\begin{itemize}[leftmargin=*]
    \item \textbf{Diversity}: We perform graph partitioning to separate the data similarity graph into segments. We treat each segment as a set of instances. We ensure the diversity of selected instances by selecting them from different segments.
    \item \textbf{Representiveness}: For each segment, we select instances with the max corresponding node degree in the data similarity graph. The selected instances thus can maximally cover the subgraph and guarantee their representativeness. 
    \item \textbf{Efficiency}: We apply a multi-level graph bisection algorithm to speed up the graph partitioning process. For the selection of each segment, we apply a simple but effective greedy algorithm. Compared to baseline methods that iteratively select over the entire graph, applying the greedy algorithm on each component can reduce the computation time.
\end{itemize}  

Compared with state-of-the-art baseline methods, our method improves the overall performance on seven datasets in three types of tasks. Besides, for all tasks, our method only needs a few seconds to complete the instance selection process. In addition, we also provide a theoretical guarantee for the effectiveness of the greedy selection algorithm.


\section{Related Work}
\mypara{In-Context Learning}
In-context learning (ICL) integrates a small number of training examples as prompts before the test input~\cite{brown2020language}, demonstrating a remarkable ability to enhance the performance of large language models (LLMs) in a wide range of downstream tasks, such as machine translation~\cite{agrawal2022context,sia2023context}, data generation~\cite{ye2022progen}, and others~\cite{wang2021want,he2023icl,panda2023differentially}. Furthermore, the advent of advanced strategies such as chain-of-thought prompting~\cite{wei2022chain} has significantly refined the efficacy of ICL, offering deeper insights and more nuanced understanding within this innovative paradigm~\cite{kim2022self,chan2022data,srivastava2022beyond,bansal2022rethinking}.

Despite its successes, ICL's efficacy is often hampered by the sensitivity to the choice of in-context examples, leading to research on optimized selection strategies~\cite{liu2021makes,lu2021fantastically,zhao2021calibrate,shi2024best}. Techniques have evolved from selecting examples close to the test input in embedding spaces~\cite{liu2021makes,wu2022self,gao2020making,rubin2021learning}. The focus has also shifted towards annotation efficiency, exploring how to find a set of examples once for all queries on the same task~\cite{zhang2023ideal,su2022selective,chang2023data}. Following existing works~\cite{zhang2023ideal,su2022selective}, we also use a graph to represent unlabeled instances and employ graph-based methods to select instances for annotation. However, our methodology distinguishes itself from existing works by focusing on efficiency in the selection of instances. As discussed in Section~\ref{sec:intro}, we aim to select diverse and representative instances while reducing the computation cost. 


\mypara{Active Learning}
Given a limited budget for labeling, active learning empowers machine learning models to achieve comparable or superior performance using a carefully selected set of labeled training instances~\cite{cohn1994improving,settles2009active,wang2023noise}. Our work on selective annotation aligns with the goal of active learning applied to graphs, specifically focusing on the selection of nodes for labeling to inform predictions~\cite{cai2017active,jia2019graph,wang2021graph,wang2022xfnc}. Traditional graph-based active learning methods employ criteria such as uncertainty~\cite{settles2008analysis} and representativeness ~\cite{li2013active} for selection. Since the ICL tasks we focus on do not involve model training or finetuning, we compare our method with basic graph active learning methods that are not incorporated with model training, like those based on node degree~\cite{cai2017active,wang2021graph} and PageRank~\cite{rodriguez2008grammar}. Our experiments indicate that while simple graph active learning methods work well in ICL, our approach still achieves better overall performance.

\vspace{-0.05in}
\section{Methods}
In this section, we will explain how to integrate a graph partition algorithm and a greedy algorithm in a selective annotation in in-context learning to reduce the annotation cost. 

\subsection{Problem Setup}
We first give the definition of the selective annotation problem. In-context learning is a paradigm that allows language models to learn tasks given only a few examples in the form of demonstration~\cite{brown2020language}. Specifically, LLMs perform in-context learning
tasks based on a task-specific prompt $\mathcal{Z}$ formed by concatenating $M$ labeled training examples $\mathcal{Z} = [ \bz_1, . . . , \bz_M ]$, where each $\bz_i$ represents one labeled example $(\bx_i, \by_i)$ consisting of the instance $\bx_i$ and label $\by_i$ (e.g., the answer of a question based on the instance). In the real world, we usually only have unlabeled samples $\mathcal{X}=\{\bx_i\}_{i=1}^{N}$, and obtaining large-scale annotated examples for ICL requires substantial manpower and financial resources.

Selective annotation aims at selecting a subset $\mathcal{L} \subset \mathcal{X}$ to be annotated, where $|\mathcal{L}| = M$ is the annotation budget, such that ICL only using prompts retrieved from the selected subset can yield good performance on the test set and thus reduce the annotation cost.

\subsection{Fast Graph-based Annotation Selection}
For the selective annotation problem, it is essential to find a subset that covers vast unlabeled data. To achieve this, we design a graph-based annotation method that balances the diversity and representativeness of the annotated samples. Briefly, we build a data similarity graph by assessing the similarity of unlabeled data embeddings. We partition the data similarity graph into distinct subgraphs by employing a multi-level graph bisection algorithm, and each subgraph is treated as a candidate set. Through a stepwise, greedy selection process of the most influential data nodes in each subgraph, we generate subsets that encapsulate the subgraph's information. Ultimately, we aggregate these subsets from all subgraphs to represent the unlabeled data. We will now provide a detailed, step-by-step explanation of the aforementioned process.

\mypara{Data Similarity Graph Generation.}
By averaging the resulting vectors over the text input words, we compute the vector representation for each unlabeled training instance using Sentence-BERT~\cite{reimers2019sentence}. We then use the embedding vectors to create the data similarity graph $\mathcal{G} = (\mathcal{V}, \mathcal{E})$ where each vertex $v_i\in \mathcal{V}$ represents an unlabeled instance $x_i\in \mathcal{X}$ as defined above. For each vertex $v\in \mathcal{V}$, we introduce edges connecting it to its $k$ nearest neighbors in terms of cosine similarity and get $\mathcal{E}$.

\begin{figure*}[!t]
\centering
\includegraphics[width=\linewidth]{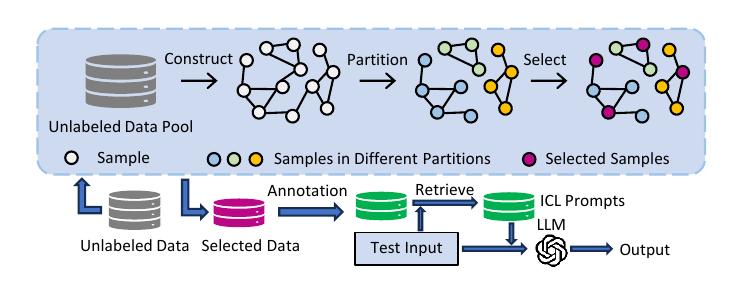}
\vspace{-0.3in}
\caption{An overview FastGAS. Given the unlabeled data pool, we initially construct a graph based on data similarity. This graph is then partitioned into distinct components. Within each component, we employ a greedy algorithm to select nodes until we reach the annotation budget. The selected instances are annotated and subsequently used to retrieve ICL prompts for the task.}
\vspace{-0.1in}
\label{fig:framework}
\end{figure*}

\mypara{Graph Partitioning.}
We aim to enhance instance \textbf{\textit{diversity}} by strategically selecting instances from various regions within the data similarity graph. The division of regions process can be conceptualized as addressing a $K$-way graph partitioning problem, where the goal is to effectively divide the graph into $K$ distinct components with approximately equal numbers of vertices but with few edges crossing between components. The formal definition of the graph partitioning problem is defined as follows: given a graph $\mathcal{G} = (\mathcal{V}, \mathcal{E})$ with $|\mathcal{V}|=N$, partition $\mathcal{V}$ into $K$ subsets, $\mathcal{V}_1, \mathcal{V}_2,..., \mathcal{V}_K$ such that $\mathcal{V}_{i} \cap \mathcal{V}_{j}= \emptyset$ for $i\neq j$, $|\mathcal{V}_{i}|$ is close $N/K$, $\bigcup_i \mathcal{V}_i=\mathcal{V}$, and the number of edges of $\mathcal{E}$ whose incident vertices belong to different subsets is minimized.

The $K$-way graph partitioning problem is an NP-complete problem. We tend to use recursive bisection to find an approximate solution with an acceptable execution time. Specifically, we first obtain a 2-way partition of $\mathcal{G}$, and then we recursively bisect the two segments independently. After log$K$ phases, $\mathcal{G}$ is partitioned into $K$ different components. Unfortunately, graph bisection is also NP-complete and has several inherent shortcomings~\cite{hendrickson1995multi}. In order to improve efficiency, we apply a multi-level graph bisection algorithm to produce high-quality partitions at low cost~\cite{karypis1998fast}. It consists of the following three phases:
\begin{itemize}[leftmargin=*]
    \item \textbf{Coarsening phase}: The graph $\mathcal{G}$ is transformed into a sequence of smaller graphs $\mathcal{G}_1, \mathcal{G}_2,..., \mathcal{G}_m$ such that $|\mathcal{V}|>|\mathcal{V}_1|>|\mathcal{V}_2|>\cdots>|\mathcal{V}_m|$.
    \item \textbf{Partitioning phase}: A 2-way partition $P_m$ of the graph $\mathcal{G}_m = (\mathcal{V}_m, \mathcal{E}_m)$ is computed that partitions $\mathcal{V}_m$ into two subgraphs, each containing half the vertices of $\mathcal{G}$.
    \item \textbf{Uncoarsening phase}: The partition $P_m$ of $\mathcal{G}_m$ is projected back to $\mathcal{G}$ by going through intermediate partitions $P_{m-1},P_{m-2},...,P_1,P_0$.
\end{itemize}

For more detailed methods of each phase, we will include them in Appendix~\ref{sec:detailpartition}.

\mypara{Greedy Node Selection.}
The graph partitioning operation divides the data similarity graph into $K$ disjoint components, ensuring diversity by treating each component as a set of instances. However, further discussion is required on the selection of suitable instances that exhibit significant \textbf{\textit{representativeness}} for each subgraph. Following the graph partitioning process, which yields $K$ subgraphs with a similar number of nodes denoted as $|\mathcal{V}_{i}| = N/K$, and considering an annotation budget $|\mathcal{L}| = M$, our objective is to choose $n = M/K$ instances within each subgraph. To mitigate potential challenges associated with high memory and computation costs, we employ a greedy selection method. In detail, for the subgraph $\mathcal{G}_i$, we first select the node $v_i^1$ that has the largest degree: $v_i^1 = \argmax_{v\in \mathcal{G}_j} d(v)$. Then, we update the subgraph $\mathcal{G}_i$ by removing the selected node $v_i^1$ and the edges connecting $v_i^1$: $\mathcal{G}_i = \mathcal{G}_i\setminus v_i^1$. The above steps are repeated $n$ times to get the selected node set $\mathcal{V}_i^{sel} = \{v_i^1,..., v_i^n\}$, and the corresponding instances $\mathcal{X}_i^{sel} = \{x_i^1,..., x_i^n\}$ are chosen to represent the instances belonging to the subgraph $\mathcal{G}_i$. The iterative form can be written as
\begin{equation}
    v_i^j=\argmax_{v\in\mathcal{G}_i\setminus\{v_i^k|k\in[1,j-1]\}} d(v)
\end{equation}
Specifically, the greedy selection algorithm guarantees the following property, demonstrating its ability to enhance the representativeness of the selected instances.
\begin{prop}\label{pro:maximize}
Given the budget $n$ and graph $\mathcal{G}$, the greedy algorithm will select $\mathcal{V}^{sel} = \{v^1,..., v^n\}$ that maximize the number of edges within $\mathcal{V}^{sel}$ and those connecting $\mathcal{V}^{sel}$ and $\mathcal{G}\setminus \mathcal{V}^{sel}$.
\begin{equation*}
\begin{aligned}
      \mathcal{V}^{sel}&=\argmax_{|\mathcal{V}|=n}|\{(u,v)|u,v\in \mathcal{V} \}| \\
    &+|\{(u,v)|u\in \mathcal{V} \hspace{0.05in}{\rm and}\hspace{0.05in} v\in \mathcal{G}\setminus \mathcal{V}\}|
\end{aligned}
\end{equation*}
\end{prop}

\begin{remark}
The greedy selection process on subgraphs can be conceptualized as a divide-and-conquer approach to selection on the entire graph, leading to improved algorithmic efficiency. Specifically, when the annotation budget $M$ is considerably smaller than the total number of nodes $N$ (i.e., $M \ll N$), the computational cost of greedy selection on the entire graph is $\mathcal{O}(2MN)$, while the cost incurred on the $K$ subgraphs is significantly reduced to $\mathcal{O}\big({\frac{2MN}{K}}\big)$.
\end{remark}


\mypara{Prompt Retrieval.}
Upon obtaining a set of instances $\mathcal{L} = \bigcup_i \mathcal{X}_i^{sel}$ through the greedy selective annotation process, we manually annotate $\mathcal{L}$ to create a comprehensive set of labeled instances. Consistent with previous studies, we leverage Sentence-BERT~\cite{reimers2019sentence} to generate embeddings for all annotated instances and identify the most similar instances to each test input based on cosine similarity.

\section{Experiment}
In this section, we evaluate the effectiveness of our method in various datasets encompassing diverse task categories. We begin by presenting the details of our experiment setups. Subsequently, the reported results demonstrate the superior performance of the proposed method in identifying an optimal selective annotation subset efficiently, outperforming established baselines. Furthermore, we conduct ablation studies to investigate the impact of crucial hyperparameters in our method.

\subsection{Setups}
\mypara{Datasets and Models}
We conduct extensive experiments in seven diverse datasets, which include six distinct tasks detailed in Table \ref{tab:dataset}.
Following existing works, each dataset adheres to the standard train/dev/test split provided by the Transformers library~\cite{wolf2019huggingface}. For datasets with publicly available test data (SST-5, XSUM, and DBpedia), we utilize the test set for evaluation. In cases where test data is not publicly accessible, consistent with previous works~\cite{zhang2023ideal, su2022selective}, we employ the dev data for evaluation. Evaluation metrics include precision for all classification and multiple-choice selection datasets and ROUGE-L~\cite{lin2004rouge} for XSUM.

Unless explicitly mentioned, we conduct all experiments using the GPT-J-6B model~\cite{gpt-j}. Additionally, we present results from tests on other models such as GPT-Neo-2.7B~\cite{black2021gpt}, OPT-6.7B~\cite{zhang2022opt}, as well as more advanced models including Llama-2-7B-Chat~\cite{touvron2023llama} and GPT-3.5-Turbo~\cite{gpt3.5turbo_doc}.

\mypara{Baselines}
In our main experiments, we conduct a comprehensive evaluation of FastGAS against random selection and two state-of-the-art selective annotation baselines: Vote-$k$~\cite{su2022selective} and IDEAL~\cite{zhang2023ideal}. Additionally, we benchmark FastGAS against other widely recognized methods in selecting core sets from extensive unlabeled data pools. These methods include (1) Top-degree~\cite{wu2019active}, which selects nodes with the largest degrees until the annotation budget is met; (2) PageRank~\cite{cai2017active}, which is used to score node representativeness in graph-based active learning; (3) Subclustering~\cite{chen2023optimization}, which initially clusters instances into $K$ groups via $K$-means, then further subdivides each into $M/K$ subclusters for centroid instance selection; and (4) Louvain~\cite{blondel2008fast}, which utilizes the Louvain community detection algorithm for graph partitioning and a greedy selection algorithm akin to FastGAS for instance selection from each community.\footnote{Since different communities contain different numbers of nodes, we select instances in proportion to the size of each community.}

To emulate real-world conditions, we follow Vote-$k$~\cite{su2022selective} and IDEAL~\cite{zhang2023ideal}, selectively annotating from a pool of 3,000 instances randomly subsampled from the original training data for each task. For robustness, we conduct this subsampling procedure three times per experiment, and the reported results represent the average across three trials. 

\mypara{Hyperparameter Setting}
In our main experiment, we create a data similarity graph for all unlabeled instances by linking each vertex to its ten nearest neighbors $(k = 10)$. For the baseline methods, we follow their hyperparameter settings to construct directed graphs~\cite{zhang2023ideal,su2022selective}. Regarding the selection of $K$ for graph partitioning, we adjust $K$ within $\{2,3,6,9\}$ for an annotation budget of $18$, and within $\{2,5,10,25,50\}$ for a budget of $100$, identifying an optimal $K$ for each task. Our method's selection time is significantly shorter than that of the baselines (Section~\ref{sec:time}), making the time spent finding the appropriate $K$ negligible. We align our annotation budgets of $18$ and $100$ with those used in Vote-$k$~\cite{su2022selective} and IDEAL~\cite{zhang2023ideal}, choosing $18$ specifically because it allows all annotated examples to fit within the context limits of LLMs without necessitating prompt retrieval. The impacts of $k$ and $K$ are detailed in Sections~\ref{sec:knn} and~\ref{sec:K}, respectively.

\subsection{Main Result}~\label{sec:mainresult}
\vspace{-0.02in}
Table~\ref{tab:main_result} presents a comparison between FastGAS and other baseline methods across annotation budgets of $|\mathcal{L}|\in\{18,100\}$. FastGAS outperforms the two existing baselines in nearly all scenarios ($13$ out of $14$). Notably, with an annotation budget of $18$, all annotated examples fit within the prompt limitations of LLMs, making the evaluation results a direct reflection of the quality of selected instances~\cite{zhang2023ideal}. When the annotation budget is $18$, FastGAS performs better than the baseline on most datasets, which shows that FastGAS can select higher-quality data. While the proposed active learning baselines outshine in specific instances (e.g., the Subclustering method excels in the DBpedia task for both annotation budgets), their general performance is hindered by their approach to balancing representativeness and diversity of the selected instances. For example, methods like Top-degree and PageRank prioritize representativeness~\cite{cai2017active}. Overall, FastGAS generally surpasses these baselines ($10$ out of $14$), demonstrating its effectiveness. Furthermore, as a deterministic selective annotation method, FastGAS operates based on a given set of unlabeled samples, mirroring the advantage seen with Vote-$k$. This means that the variability in FastGAS's performance stems exclusively from the manner in which unlabeled samples are gathered. This significantly enhances the robustness of ICL by ensuring consistency in the selection process~\cite{su2022selective}. We provide detailed results that contain the maximum and minimum values of each task in Appendix~\ref{sec:detailresult}.

\begin{table*}
\caption{The results of FastGAS and six baselines on seven distinct datasets with annotation budgets of $100$ and $18$, utilizing similarity-based prompt retrieval for all methods. We report the average results with three different runs for each method. 
Bold fonts indicate the best performance, while underlines denote the second-best results.}
\centering
\begin{tabular}{c c c c c c c c c }
\toprule[1.2pt]
    \multirow{2}{*}{\textbf{$|\mathcal{L}|$}} &\multirow{2}{*}{\textbf{Methods}} & \multicolumn{5}{c}{\textbf{Classification}} & \textbf{Multi-Choice}  &\textbf{Generation} \\ \cmidrule{3-7} 
     &     &MRPC &SST-5 &MNLI &DBpedia &RTE &HellaSwag &XSUM  \\ 
    \midrule
    100 &Random  &64.32 &49.61 &38.15 &89.84 &55.34 &64.71 &17.24\\
    100 &Vote-$k$ &64.60 &46.61 &38.93 &89.19 &57.68 &65.89 &19.55\\
    100 &IDEAL  &\underline{65.49}  &\underline{49.87} &41.02  &90.63  &\underline{58.98} &64.97 &\underline{19.68}\\
    \hdashline
    100 &Top-degree &62.76 &42.32 &41.02 &83.85 &51.69 &\underline{66.67} &19.39\\
    100 &PageRank  &64.84 &40.71 &\textbf{44.17} &83.72 &53.38 &66.01 &19.55 \\
    100 &Subclustering &64.84 &49.48 &41.28 &\textbf{92.32} &57.55 &65.62 &19.18 \\
    100 &Louvain  &59.63 &39.58 &41.14 &86.33 &53.52 &65.89 &19.27\\
    100 &\textbf{FastGAS} &\textbf{66.15}  &\textbf{50.26} &\underline{42.06}  &\underline{90.76}  &\textbf{61.98} &\textbf{67.45} &\textbf{20.00}\\ 
    \midrule[1.1pt]
    18 &Random  &55.47 &42.97 &37.76 &83.20 &54.95 &65.49 &16.63\\
    18 &Vote-$k$ &56.90  &41.78  &39.45  &88.02 &\textbf{56.64} &\underline{66.02} &19.45  \\
    18 &IDEAL  &\underline{63.80}   &44.92 &39.58 &83.20   &53.65 &65.89 &19.21\\ 
    \hdashline
    18 &Top-degree &48.57 &39.45 &\underline{41.93} &77.34 &54.43 &65.76 &20.05\\
    18 &PageRank  &46.09 &39.71 &40.49 &79.04 &56.07 &65.89 &19.82\\
    18 &Subclustering &61.46 &\underline{46.05} &37.63 &\textbf{89.06} &\underline{56.63} &65.76 &19.49\\
    18 &Louvain  &61.33 &38.02 &38.02 &83.59 &55.86 &65.62 &\underline{20.23}\\
    18 &\textbf{FastGAS} &\textbf{65.23}   &\textbf{46.61} &\textbf{44.53} &\underline{88.93} &56.51 &\textbf{66.67} &\textbf{20.26}\\
\bottomrule[1.2pt]
\end{tabular}
\vspace{-0.1in}
\label{tab:main_result}
\end{table*}
\subsection{Effect of $k$}\label{sec:knn}

\begin{figure}[htbp]
\centering
\includegraphics[width=\linewidth]{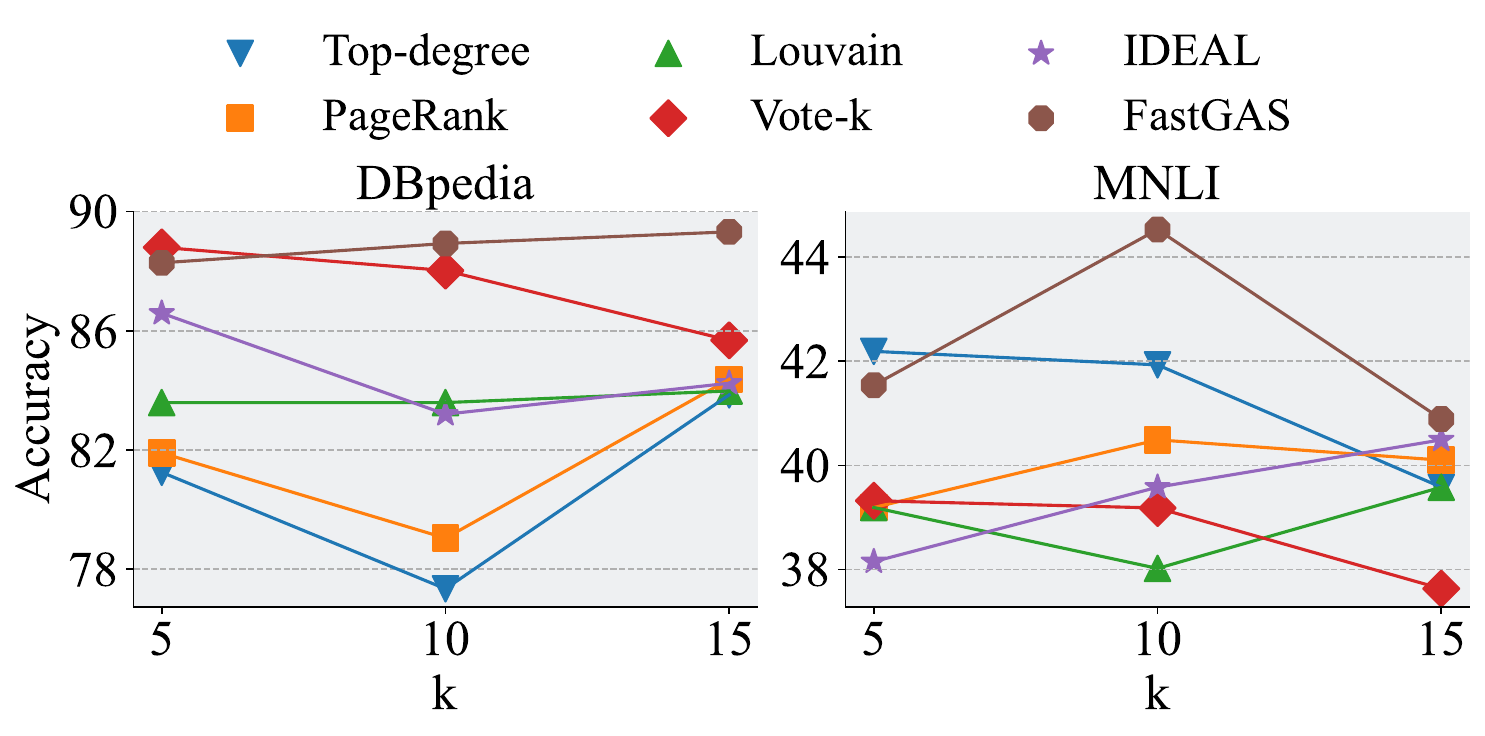}
\vspace{-0.3in}
\caption{Comparison of our method and other graph-based baselines with respect to different $k$ for the construction of the graph. }
\label{fig:diff_k}
\vspace{-0.2in}
\end{figure}

We compare FastGAS and other graph-based baselines with respect to different $k$ for the construction of the graph, and the result is shown in Figure~\ref{fig:diff_k}. Initially, our method consistently exceeds other graph-based baselines in various settings. Second, we observe that the ideal $k$ value varies among different methods and datasets. Specifically, for the MNLI dataset, most methods perform better with $k=10$.
We also conclude that, for FedGAS, $k$ must be carefully balanced, neither too large nor too small. A smaller $k$ value, as well as a larger one, impedes the ability to differentiate between distinct representative nodes in the graph, complicating the selection process by potentially choosing multiple similar nodes. Concurrently, a larger $k$ value could also increase the computational time.

\subsection{Evaluating Time Efficiency}\label{sec:time}
\begin{figure}[htbp]
\centering
\includegraphics[width=\linewidth]{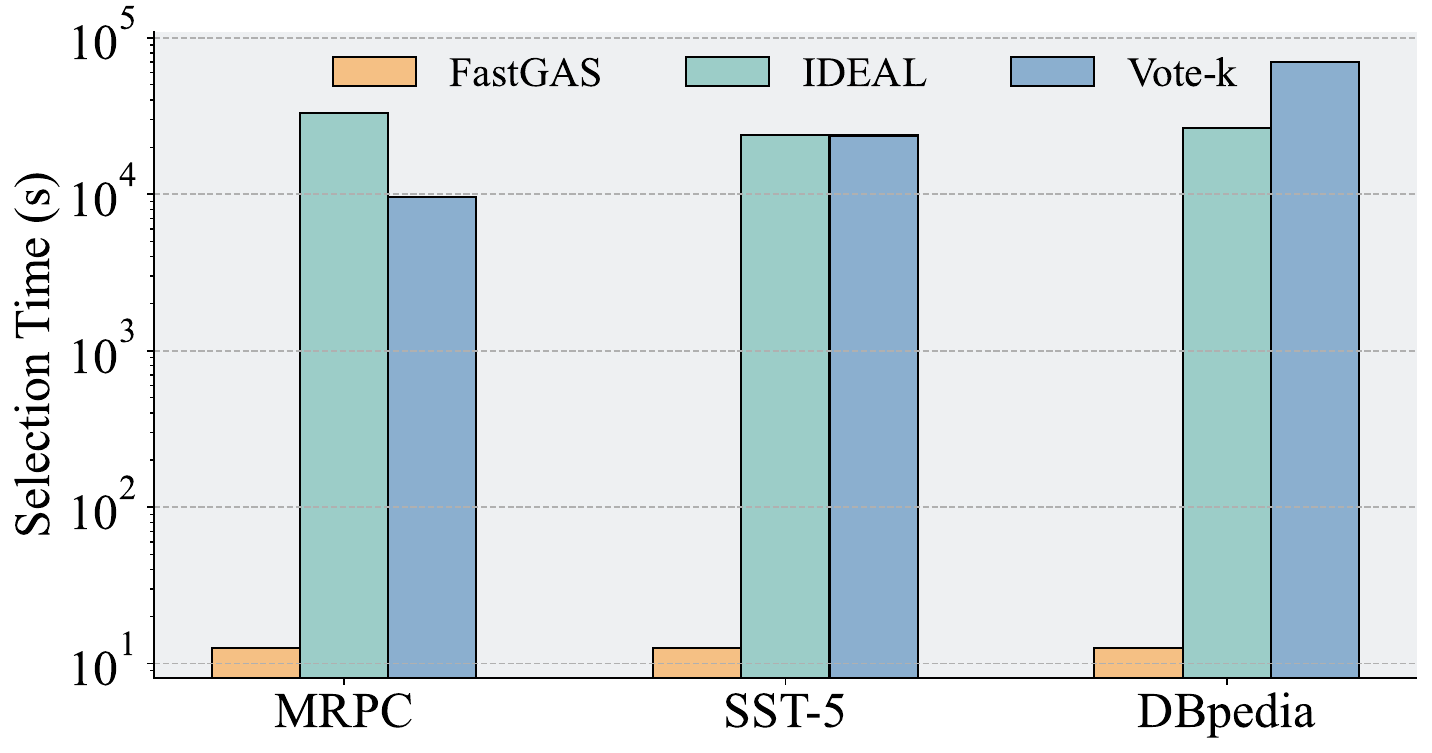}
\caption{Comparison of our method and two baselines on three classification tasks (MPRC, SST-5, and DBpedia) with respect to time consumption during subset selection. The annotation budget is $100$. The y-axis represents the time consumption with a log scale.}
\label{fig:time_100}
\vspace{-0.2in}
\end{figure}

In Section~\ref{sec:intro}, we illustrate that FastGAS drastically reduces the time cost compared to two existing methods with an annotation budget of $18$. Figure~\ref{fig:time_100} expands on this by showing the time consumption of our method and two baselines for an annotation budget of $100$. Relative to Figure~\ref{fig:time_18}, the time needed for FastGAS does not increase much. This is because the most time-intensive processes in FastGAS, i.e., constructing and partitioning the data similarity graph, are not affected by the annotation budget. In contrast, the time costs of the two baselines, especially IDEAL, increase sharply; for nearly all tasks, IDEAL and Vote-$k$ require more than eight hours for selection. As the annotation budget increases, the efficiency of FastGAS in example selection becomes even more pronounced.

\subsection{Effect of Number of Partitions}\label{sec:K}
\begin{figure}[htbp]
\centering
\includegraphics[width=\linewidth]{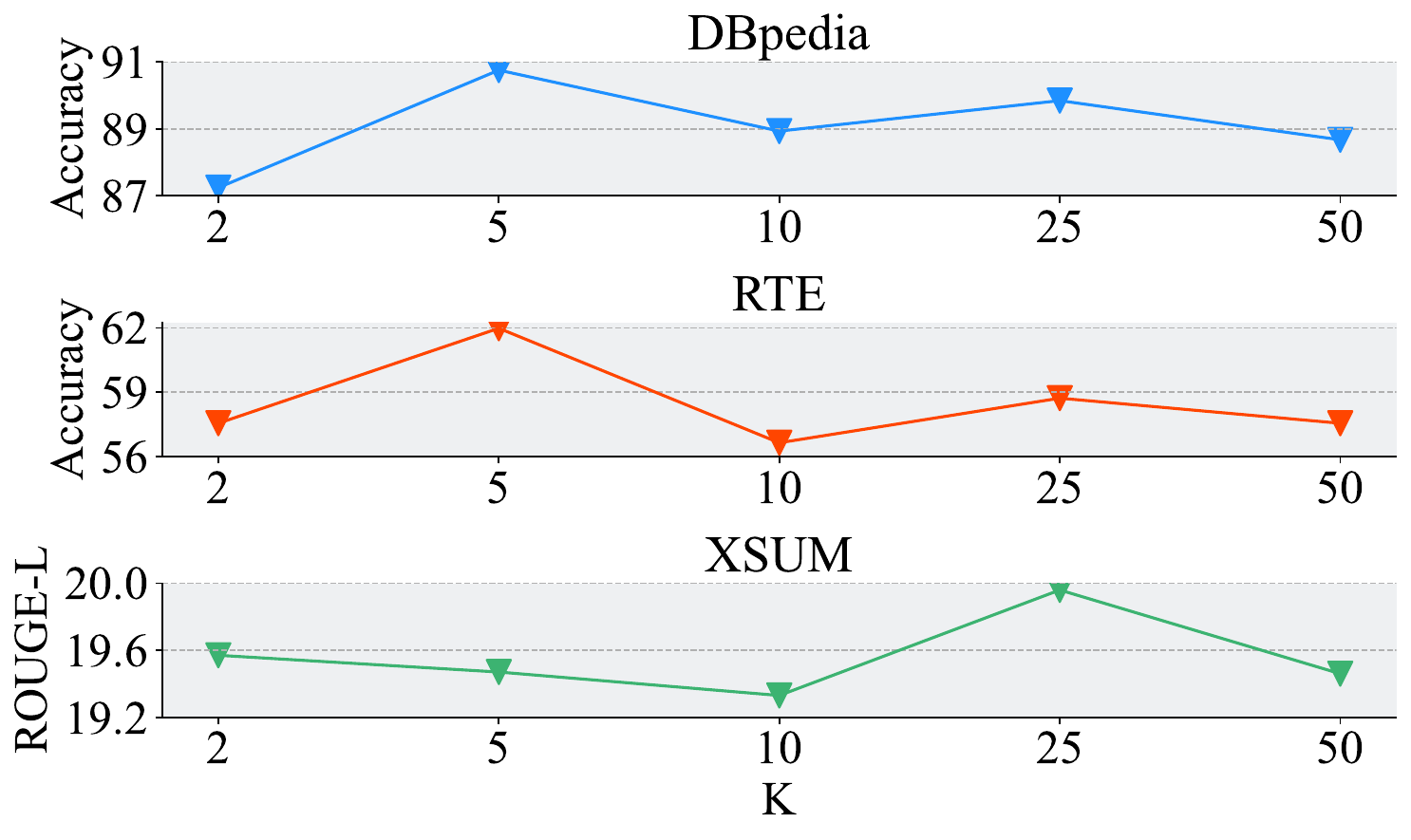}
\vspace{-0.3in}
\caption{Performance of FastGAS across different numbers of partitions with an annotation budget of $100$.}
\vspace{-0.3in}
\label{fig:diff_commun}
\end{figure}

The hyperparameter $K$ plays a critical role in graph partitioning, determining the number of components into which the graph is divided. We explore $K$'s impact on FastGAS and offer insights for adjusting it. Figure~\ref{fig:diff_commun} illustrates FastGAS's performance in three datasets with an annotation budget of $100$. The figure demonstrates that enhancing $K$ up to a certain point improves FastGAS's performance; however, the increase beyond this optimal threshold does not result in further performance gains. A small $K$ results in relatively rough partitioning, such that significantly different data points are grouped into the same subgraph. Selecting through coarsely partitioned subgraphs can obscure meaningful distinctions in the data, resulting in the loss of the diversity of the selected nodes. Conversely, with a sufficiently large $K$, the graph partitioning algorithm results in more edges being cut during the partitioning process, thus affecting the degree of nodes in each disjoint subgraph. As a result, nodes with a large degree in the original graph may lose a large number of edges due to partitioning and thus be ignored by the greedy algorithm. This alteration hampers the greedy algorithm's ability to select representative nodes effectively. Hence, a balanced $K$ value facilitates the optimal performance of FastGAS by ensuring a balance of representativeness and diversity.


\subsection{Random Prompt Retrieval}\label{sec:randomretr}
In Section~\ref{sec:mainresult}, we evaluate the performance of FastGAS against other baselines using a similarity-based prompt retrieval method. Building on previous research~\cite{zhang2023ideal,su2022selective}, we investigate the impact of employing a random retrieval method, specifically, by randomly selecting labeled instances as prompts for each test instance under the annotation budget $100$ and $18$. The findings are presented in Table~\ref{tab:randomretr}. We note a marked decline in the effectiveness of selective annotation methods under large annotation budgets when prompts are chosen through random selection. This deterioration may stem from the increased likelihood of selecting instances from the larger labeled pool that are less relevant to the test sample, as determined by their distance in the embedding space~\cite{liu2021makes}. Significantly, FastGAS continues to outperform the two baseline methods across all datasets, even when employing random retrieval. With an annotation budget of $18$, all annotated instances fit within the prompt capacity of LLMs, making the order of instances the only difference between the two prompt retrieval methods. The performance of FastGAS underscores its capability to produce a more reliable training subset for ICL tasks~\cite{chang2023data}.
\begin{table}
\small
\tabcolsep = 3pt
\centering
\caption{Comparison of random and similar prompt retrieval. The selection approach, when paired with a similarity-based prompt retrieval method, generally outperforms its counterpart, which utilizes a random prompt retrieval technique.}
\begin{tabular}{c c c| c c c}
    \toprule[1.2pt]
    \multicolumn{3}{c|}{\textbf{Methods}} & \multicolumn{3}{c}{\textbf{Datasets}}\\ \cmidrule{1-6}
    $|\mathcal{L}|$ &Selection &Retrieval &MRPC &MNLI &HellaSwag \\
    \midrule[1.1pt]
    100 &Vote-$k$ &Similar &64.60 &38.93 &65.89 \\
    100 &IDEAL &Similar &65.49 &41.02 &64.97\\
    100 &FastGAS &Similar &\textbf{66.15} &\textbf{42.06} &\textbf{67.45}\\
    \midrule
    100 &Vote-$k$ &Random &60.67 &37.76 &64.58\\
    100 &IDEAL &Random &62.50 &39.06 &66.80\\
    100 &FastGAS &Random &\textbf{62.50} &\textbf{40.88} &\textbf{67.32}\\
    \midrule[0.8pt]
    18 &Vote-$k$ &Similar &56.90 &39.45 &66.02 \\
    18 &IDEAL &Similar &63.80 &39.58 &65.89\\
    18 &FastGAS &Similar &\textbf{65.23} &\textbf{44.53} &\textbf{66.67}\\
    \midrule
    18 &Vote-$k$ &Random &54.56 &41.02 &66.54\\
    18 &IDEAL &Random &64.19 &38.15 &66.54\\
    18 &FastGAS &Random &\textbf{64.71} &\textbf{44.01} &\textbf{67.19}\\
    \bottomrule[1.2pt]
\end{tabular}
\vspace{-0.1in}
\label{tab:randomretr}
\end{table}

\subsection{Evaluation on Different Language Models}

We conduct evaluations of FastGAS on various language models, including GPT-Neo 2.7B~\cite{black2021gpt}, OPT-6.7B~\cite{zhang2022opt}, Llama-2-7B-Chat~\cite{touvron2023llama}, and GPT-3.5-Turbo~\cite{gpt3.5turbo_doc}\footnote{In a recent update, OpenAI announced the deprecation of the logprobs endpoint. Consequently, in our experiment, we employ the Vote-$k$ method, specifically `Fast vote-$k$,' which does not depend on this value.}, applying the same instructions across each dataset. Table~\ref{tab:gpj_opt} shows the performance of three selective annotation methods on smaller language models (GPT-Neo 2.7B and OPT-6.7B) across three datasets, where FastGAS overall surpasses two baseline annotation methods. Notably, OPT-6.7B exhibits lower performance on the XSUM summarization task compared to other LLMs, a finding echoed in~\cite{tam2022evaluating}. However, FastGAS remains superior to the baselines even when using the OPT-6.7B model.

\begin{figure}[htbp]
\centering
\includegraphics[width=\linewidth]{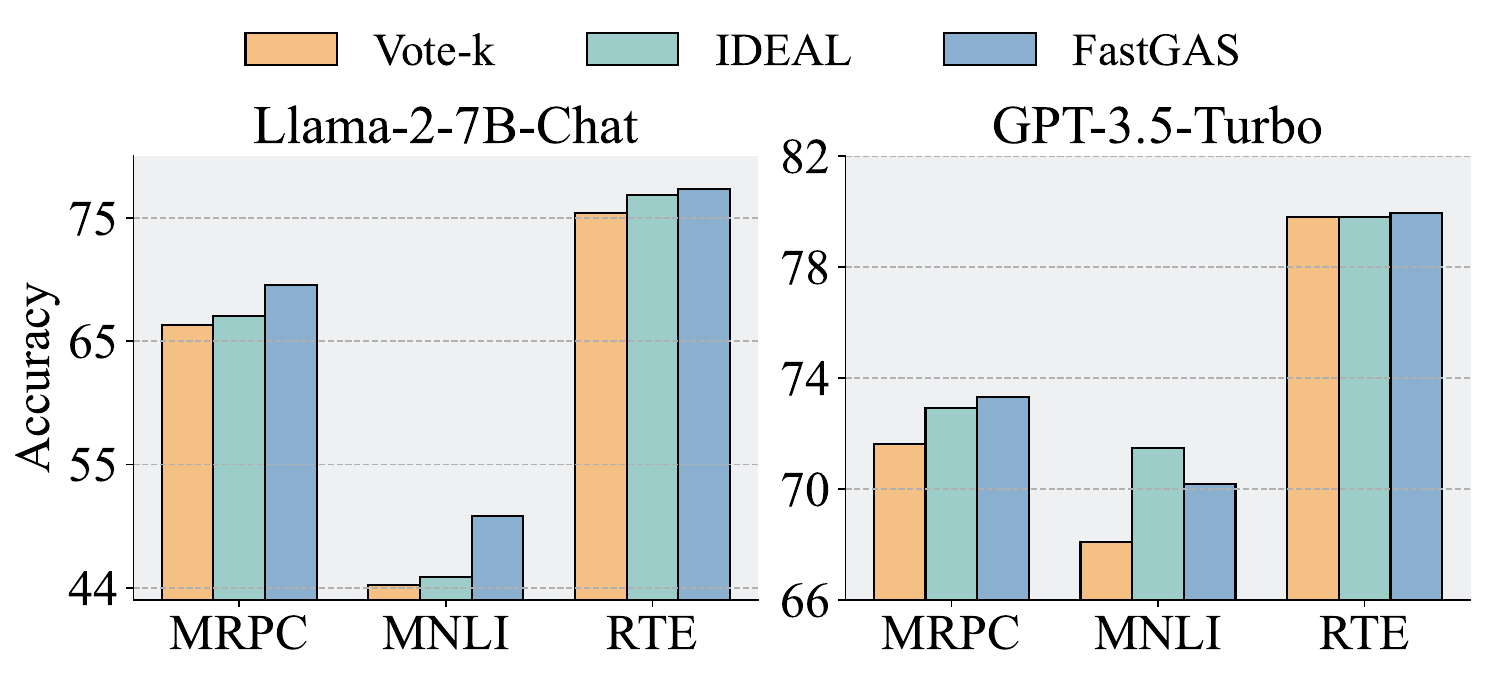}
\caption{Comparative performance of FastGAS versus baselines using Llama-2-7B-Chat and GPT-3.5-Turbo models across various datasets with an annotation budget of $18$. 
}
\label{fig:gpt_llama}
\end{figure}

\begin{table}
\small
\tabcolsep = 3.5pt
\centering
\caption{Comparative performance of FastGAS versus baselines using GPT-Neo-2.7B and OPT-6.7B models across various datasets with an annotation budget of $18$. FastGAS generally outperforms baseline methods across a variety of models and datasets.}
\begin{tabular}{c c| c c c}
    \toprule[1.2pt]
    \multicolumn{2}{c}{\textbf{Methods}} & \multicolumn{3}{c}{\textbf{Datasets}}\\ \cmidrule{1-5}
    Selection &Model &MRPC &DBpedia &XSUM \\
    \midrule[1.1pt]
    Vote-$k$ &GPT-Neo-2.7B &57.42 &80.73 &19.45 \\
    IDEAL &GPT-Neo-2.7B &65.89 &78.38 &19.69\\
    FastGAS &GPT-Neo-2.7B &\textbf{66.02} &\textbf{80.86} &\textbf{20.15}\\
    \midrule[0.8pt]
    Vote-$k$ &OPT-6.7B &36.59 &88.02 &6.86 \\
    IDEAL &OPT-6.7B &\textbf{45.18} &83.20 &6.13\\
    FastGAS &OPT-6.7B &44.66 &\textbf{88.93} &\textbf{6.89}\\
    \bottomrule[1.2pt]
\end{tabular}
\vspace{-0.05in}
\label{tab:gpj_opt}
\end{table}
Further experiments on more advanced LLMs are presented in Figure~\ref{fig:gpt_llama}, confirming FastGAS's enhanced performance with larger language models. This underscores FastGAS as a versatile approach effective across LLMs of varying sizes. Particularly, the performance leap with GPT-3.5-Turbo, especially notable on the MNLI dataset (improving from 50.18\% with Llama-2-7B-Chat to 70.18\% with GPT-3.5-Turbo), highlights the advantage of larger models equipped with more parameters and extensive training data, thereby bolstering their capability in classification tasks.

\vspace{-0.1in}
\section{Conclusion}

Recent advancements have showcased the capability of large language models (LLMs) to adapt to new tasks with just a few demonstration instances. While existing approaches aim to enhance model performance by selecting a limited number of instances to annotate for prompts, they are hindered by significant computational demands and lengthy processing times. To address these challenges, we introduce a graph-based selection algorithm, FastGAS, accompanied by a theoretical analysis of its effectiveness. Our empirical evaluations reveal that this method outperforms others in seven distinct tasks, demonstrating exceptional efficacy. Unlike conventional methods that require high computation costs, our method greatly improves the efficiency of selecting instances, substantially enhancing its applicability to practical scenarios. Additionally, we demonstrate FastGAS's versatility across LLMs of various sizes. We hope that these insights will help researchers in devising effective ICL strategies to optimize LLM performance.

\section{Limitation}
The primary constraint of our study is the inability to automatically select the most appropriate number of partitions ($K$) and the most appropriate number of neighbors ($k$) during the data similarity graph construction. Given the relatively short execution time of our method, conducting multiple trials to identify the ideal $K$ and $k$ value is viable. However, the cost of the inference phase could restrict the feasibility of such extensive experimentation in practical settings. To enhance efficiency, FastGAS adopts a greedy selection process that is carried out separately for each piece. However, we have not explored how the interrelations between samples across different graph pieces influence the overall instance selection. Additionally, due to hardware limitations and available time, our research only covered LLMs up to 7B in size. Future investigations will aim to assess FastGAS's efficacy with larger LLMs and across a broader array of tasks.

\section{Ethics Statement}
This work introduces FastGAS, a graph-based selective annotation method that can effectively and efficiently select high-quality instances for in-context learning tasks. While acknowledging the need for responsible usage of the proposed method, we do not foresee major negative societal impacts.

\section{Acknowledgements}
This work was supported in part by the US National Science Foundation (NSF) under awards ECCS-2332060, CPS-2313110, ECCS-2143559, ECCS-2033671, SII-2132700, IIS-2006844, IIS-2144209, IIS-2223769, CNS2154962, and BCS-2228534, and the Commonwealth Cyber Initiative under awards VV-1Q23-007, HV-2Q23-003, and VV-1Q24-011.

\bibliography{acm}
\bibliographystyle{acl_natbib}

\appendix
\onecolumn

\section{General experimental conditions}
Our implementation of FastGAS and baselines is primarily conducted using PyTorch~\cite{paszke2019pytorch}. For the GPT-3.5-Turbo experiments, we utilize the OpenAI API. The models GPT-J-6B, GPT-Neo 2.7B, Llama-2-7B-Chat, and OPT-6.7B are sourced from the Huggingface Transformers Library~\cite{wolf2019huggingface}. All experiments are performed on a single NVIDIA RTX A6000 GPU with 48GB of memory.

\section{The Detailed Information of Seven Datasets}
\begin{table*}[h]
\small
\centering
\caption{The detailed information of seven datasets}
\begin{tabular}{c c c }
    \toprule
             & \textbf{Dataset} & \textbf{Task}\\
        \midrule
        \multirow{2}{*}{\textbf{Classification}}   &MRPC~\cite{dolan2004unsupervised} &Paraphrase Detection \\
            &SST-5~\cite{socher2013recursive} &Sentiment Analysis \\
            &DBpedia~\cite{lehmann2015dbpedia} &Topic Classification \\
            &MNLI~\cite{williams2018broad} &Natural Language Inference \\
            &RTE~\cite{bentivogli2009fifth} &Natural Language Inference \\
        \midrule
        \textbf{Multiple-Choice} &HellaSwag~\cite{zellers2019hellaswag} &Commonsense Reasoning \\
        \midrule
        \textbf{Generation} &XSUM~\cite{narayan2018don} &Summarization\\
    \bottomrule
\end{tabular}
\label{tab:dataset}
\end{table*}

\section{Detailed Graph Partition Algorithm}\label{sec:detailpartition}
\begin{itemize}
    \item \textbf{Coarsening phase}: The graph $\mathcal{G}$ is transformed into a sequence of smaller graphs $\mathcal{G}_1, \mathcal{G}_2,..., \mathcal{G}_m$ such that $|\mathcal{V}|>|\mathcal{V}_1|>|\mathcal{V}_2|>\cdots>|\mathcal{V}_m|$.
    \item \textbf{Partitioning phase}: A 2-way partition $P_m$ of the graph $\mathcal{G}_m = (\mathcal{V}_m, \mathcal{E}_m)$ is computed that partitions $\mathcal{V}_m$ into two parts, each containing half the vertices of $\mathcal{G}$.
    \item \textbf{Uncoarsening phase}: The partition $P_m$ of $\mathcal{G}_m$ is projected back to $\mathcal{G}$ by going through intermediate partitions $P_{m-1},P_{m-2},...,P_1,P_0$
\end{itemize}

\mypara{Coarsening phase}
Due to the construction of the data similarity graph, the degree of most vertices is close to the graph's average degree. To generate coarser graphs, we employ a strategy of finding a random matching and merging the matched vertices into a multi-node~\cite{bui1993heuristic,hendrickson1995multi}. A graph's matching is a set of edges, each pair of which shares no common vertex. We create a subsequent coarser graph, $\mathcal{G}_{i+1}$, from $\mathcal{G}_{i}$ by matching in $\mathcal{G}_{i}$ and merging the matched vertices. We utilize a random algorithm akin to those detailed in~\cite{bui1993heuristic,hendrickson1995multi}.

The algorithm operates as follows: vertices are processed in a random sequence. For an unmatched vertex $u$, an unmatched adjacent vertex $v$ is randomly chosen, and the edge $(u,v)$ is added to the matching. If there is no unmatched adjacent vertex for $u$, then $u$ remains unmatched in this random matching process. The computational complexity of this algorithm is $\mathcal{O}(|\mathcal{E}|)$.

\mypara{Partitioning phase}
To efficiently bisect the graph, we simply initiate from a vertex and expand a region around it using a breadth-first approach until half of the vertices are encompassed~\cite{ciarlet1996validity,goehring1994heuristic}. Given that the bisection quality is highly dependent on the initial vertex selection~\cite{karypis1998fast}, we employ a strategy similar to~\cite{karypis1998fast}, wherein we randomly select 10 vertices and subsequently expand 10 distinct regions from each. The trail yielding the smallest edge cut is then chosen for the partition.

\mypara{Uncoarsening phase}
During the uncoarsening phase, as each vertex $u \in \mathcal{G}_{i+1}$ represents a unique subset $\mathcal{U}$ of vertices from $\mathcal{G}_{i}$, we derive partition $P_{i}$ from $P_{i+1}$ by assigning the vertices in $\mathcal{U}$ to the same part that vertex $u$ belongs to. Although $P_{i+1}$ may represent a local minimum partition for $\mathcal{G}_{i+1}$, the corresponding partition 
$P_{i}$ might not be at a local minimum with respect to $\mathcal{G}_{i}$. To refine this uncoarsened partition, we employ the KL algorithm~\cite{kernighan1970efficient}, which iteratively searches for and swaps subsets of vertices between partitions to achieve a lower edge cut. This swapping process continues until no further improvements can be made, indicating that the partition has reached a local minimum.

\section{Proof of Proposition~\ref{pro:maximize}}
\begin{proof}
We employ an inductive approach to demonstrate that the node set $\mathcal{V}^{sel}$, selected by the greedy algorithm, satisfies the maximum cover property,
\begin{equation}
      \mathcal{V}^{sel}=\argmax_{|\mathcal{V}|=n}|\{(u,v)|u,v\in \mathcal{V}\}|+|\{(u,v)|u\in \mathcal{V} \hspace{0.05in}{\rm and}\hspace{0.05in} v\in \mathcal{G}\setminus \mathcal{V}\}|\label{eq:pf1}
\end{equation}
When $n=1$, $\mathcal{V}^{sel} = \{v^1\}$. Thus,
\begin{equation*}
    |\{(u,v)|u\in \mathcal{V} \hspace{0.05in}{\rm and}\hspace{0.05in} v\in \mathcal{G}\setminus \mathcal{V}\}| = {\rm degree} (v^1)\hspace{0.05in}{\rm and}\hspace{0.05in}|\{(u,v)|u,v\in \mathcal{V} \}|=0.
\end{equation*}
By selecting $v_1=\argmax_{v\in\mathcal{G}}d(v)$, the maximum cover property holds for $n=1$.

\noindent Assume it holds for $n=k$, let $\mathcal{V}_k^{sel}$ denotes the selected node set at the budget $k$ and $\mathcal{V}_{k+1}^{sel}=\mathcal{V}_k^{sel}\cup\{v^{k+1}\}$. For the first term in Eq.~\ref{eq:pf1}, we have
\begin{equation}\label{eq:first}
    |\{(u,v)|u,v\in \mathcal{V}_{k+1}^{sel}\}|=|\{(u,v)|u,v\in \mathcal{V}_k^{sel}\}|+|\{(u,v^{k+1})|u\in \mathcal{V}_k^{sel}\}|.
\end{equation}
\noindent For the second term in Eq.~\ref{eq:pf1}, we have
\begin{equation}\label{eq:second}
\begin{aligned}
    &|\{(u,v)|u\in \mathcal{V} _{k+1}^{sel}\hspace{0.05in}{\rm and}\hspace{0.05in} v\in \mathcal{G}\setminus \mathcal{V}_{k+1}^{sel}\}|\\
    =&|\{(u,v)|u\in \mathcal{V} _{k}^{sel}\hspace{0.05in}{\rm and}\hspace{0.05in} v\in \mathcal{G}\setminus \mathcal{V}_{k+1}^{sel}\}|+|\{(v^{k+1},v)|\hspace{0.05in} v\in \mathcal{G}\setminus \mathcal{V}_{k+1}^{sel}\}|\\
    =&|\{(u,v)|u\in \mathcal{V} _{k}^{sel}\hspace{0.05in}{\rm and}\hspace{0.05in} v\in \mathcal{G}\setminus \mathcal{V}_{k+1}^{sel}\}|+|\{(v^{k+1},v)|\hspace{0.05in} v\in \mathcal{G}\setminus \mathcal{V}_{k}^{sel}\}|\\
    =&|\{(u,v)|u\in \mathcal{V} _{k}^{sel}\hspace{0.05in}{\rm and}\hspace{0.05in} v\in \mathcal{G}\setminus \mathcal{V}_{k}^{sel}\}|-|\{(u,v^{k+1})|u\in \mathcal{V} _{k}^{sel}\}|+|\{(v^{k+1},v)|\hspace{0.05in} v\in \mathcal{G}\setminus \mathcal{V}_{k}^{sel}\}|.\\
\end{aligned}    
\end{equation}
The second equality sign holds because
\begin{equation*}
\begin{aligned}
    &|\{(v^{k+1},v)|\hspace{0.05in} v\in \mathcal{G}\setminus \mathcal{V}_{k+1}^{sel}\}|=|\{(v^{k+1},v)|\hspace{0.05in} v\in \mathcal{G}\setminus \mathcal{V}_{k}^{sel}\}|-|\{(v^{k+1},v)|\hspace{0.05in} v\in \mathcal{V}_{k+1}^{sel}\setminus \mathcal{V}_{k}^{sel}\}|,\\
    &|\{(v^{k+1},v)|\hspace{0.05in} v\in \mathcal{V}_{k+1}^{sel}\setminus \mathcal{V}_{k}^{sel}\}|=|\{(v^{k+1},v)|\hspace{0.05in} v\in \{v^{k+1}\}\}| = 0.
\end{aligned}
\end{equation*}
\noindent We then combine Eq.~\ref{eq:first} and Eq.~\ref{eq:second} and obtain that
\begin{equation}
\begin{aligned}
    &|\{(u,v)|u,v\in \mathcal{V}_{k+1}^{sel}\}|+|\{(u,v)|u\in \mathcal{V} _{k+1}^{sel}\hspace{0.05in}{\rm and}\hspace{0.05in} v\in \mathcal{G}\setminus \mathcal{V}_{k+1}^{sel}\}|\\
    =&|\{(u,v)|u,v\in \mathcal{V}_k^{sel}\}|+|\{(u,v^{k+1})|u\in \mathcal{V}_k^{sel}\}|+|\{(u,v)|u\in \mathcal{V} _{k}^{sel}\hspace{0.05in}{\rm and}\hspace{0.05in} v\in \mathcal{G}\setminus \mathcal{V}_{k}^{sel}\}|\\
    &-|\{(u,v^{k+1})|u\in \mathcal{V} _{k}^{sel}\}|+|\{(v^{k+1},v)|\hspace{0.05in} v\in \mathcal{G}\setminus \mathcal{V}_{k}^{sel}\}|\\
    =&|\{(u,v)|u,v\in \mathcal{V}_k^{sel}\}|+|\{(u,v)|u\in \mathcal{V} _{k}^{sel}\hspace{0.05in}{\rm and}\hspace{0.05in} v\in \mathcal{G}\setminus \mathcal{V}_{k}^{sel}\}|+|\{(v^{k+1},v)|\hspace{0.05in} v\in \mathcal{G}\setminus \mathcal{V}_{k}^{sel}\}|
\end{aligned}
\end{equation}
\noindent Since $\mathcal{V}_k^{sel}$ satisfies that
\begin{equation*}
      \mathcal{V}_k^{sel}=\argmax_{|\mathcal{V}|=k}|\{(u,v)|u,v\in \mathcal{V}\}|+|\{(u,v)|u\in \mathcal{V} \hspace{0.05in}{\rm and}\hspace{0.05in} v\in \mathcal{G}\setminus \mathcal{V}\}|,
\end{equation*}
and from the algorithm, we have
\begin{equation*}
    v^{k+1}=\argmax_{v\in\mathcal{G}\setminus\{v_i|i\in[1,k]\}} {\rm degree}(v) = \argmax_{v\in\mathcal{G}\setminus \mathcal{V}_{k}^{sel}} |\{(v,u)|\hspace{0.05in} u\in \mathcal{G}\setminus \mathcal{V}_{k}^{sel}\}|.
\end{equation*}
\noindent Thus
\begin{equation*}
\begin{aligned}
    \mathcal{V}_{k+1}^{sel}&=\mathcal{V}_k^{sel}\cup\{v^{k+1}\}\\
    &=\argmax_{|\mathcal{V}|=k+1}|\{(u,v)|u,v\in \mathcal{V}\}|+|\{(u,v)|u\in \mathcal{V} \hspace{0.05in}{\rm and}\hspace{0.05in} v\in \mathcal{G}\setminus \mathcal{V}\}|.
\end{aligned}
\end{equation*}
\noindent In conclusion, given the budget $n$ and graph $\mathcal{G}$, the $\mathcal{V}^{sel}$ returned by the greedy algorithm satisfies the property in Eq.\ref{eq:pf1}. 
\end{proof}

\section{Example Prompt of Each Task}
\subsection{MRPC}
\mypara{Input:} Are the following two sentences `equivalent' or `not equivalent'?\\
Excluding autos, retail sales rose by 0.3 \% in September, lower than a forecast rise of 0.5 \% ..\\
Retail sales fell 0.2 percent overall in September compared to forecasts of a 0.1 percent dip ..\\
answer:\\
\mypara{Output:} not equivalent
\subsection{SST-5}
\mypara{Input:} How do you feel about the following sentence?\\
... the film's considered approach to its subject matter is too calm and thoughtful for agitprop, and the thinness of its characterizations makes it a failure as straight drama.\\
answer:\\
\mypara{Output:} very negative
\subsection{MNLI}
\mypara{Input:} Quite an hour, or even more, had elapsed between the time when she had heard the voices and 5 o'clock when she had taken tea to her mistress. Based on that information, is the claim A day had passed from when she'd taken the tea in. "True," "False," or "Inconclusive"?\\
answer:\\
\mypara{Output:} False
\subsection{DBpedia}
\mypara{Input:} title: Grace Evangelical Lutheran Church (Minneapolis Minnesota); content:  Grace Evangelical Lutheran Church is a church in Minneapolis, Minnesota, United States, adjacent to the University of Minnesota East Bank campus. The church was built in 1915-1917 by a Swedish Lutheran congregation to serve university students. It was designed by Chapman and Magney and built in the Gothic Revival style. The congregation was organized in Minneapolis in 1903 by the Swedish immigrant-dominated Augustana Evangelical Lutheran Church.\\
\mypara{Output:} building
\subsection{RTE}
\mypara{Input:} He met U.S. President, George W. Bush, in Washington and British Prime Minister, Tony Blair, in London..\\
question: Washington is part of London. True or False?\\
answer:\\
\mypara{Output:} False
\subsection{Hellaswag}
\mypara{Input:} The topic is Cleaning sink. A middle aged female talks about a cleaning product. The female opens a container of cleaner and puts it on a rag. the female,\\
\mypara{Options:}  ["then inflames a different cleaner to clean a sock.",
                "uses the rag to spray down a wall.",
                "washes the rug thoroughly and scratches it.",
                "then uses the rag to rub the inside of the sink."]\\
\mypara{Output:} then uses the rag to rub the inside of the sink.
\subsection{Xsum}
\mypara{Input:} Stead curled home with 14 minutes remaining to cap a fine comeback at the Northern Gas and Power Stadium after Louis Lang cancelled out Toto Nsiala's opener. Hartlepool flew out of the blocks and took an eighth-minute lead when Nsiala bundled home at the back post after Lewis Alessandra beat his man and sent in a pin-point cross.\\
......\\
Substitution, Notts County. Vadaine Oliver replaces Jonathan Forte because of an injury. Attempt saved. Nathan Thomas (Hartlepool United) right footed shot from the left side of the box is saved in the bottom right corner. Corner,  Hartlepool United. Conceded by Matt Tootle. Attempt missed. Michael Woods (Hartlepool United) right footed shot from outside the box is close, but misses the top right corner. Foul by Billy Paynter (Hartlepool United). Stanley Aborah (Notts County) wins a free kick in the attacking half...\\
\mypara{Output:} Jon Stead struck the winner as Notts County came from behind to earn victory at Hartlepool United in League Two.

\section{Detailed Main Results}\label{sec:detailresult}
In Section~\ref{sec:mainresult}, we present the average evaluation outcomes of various methods across three random trials. This section offers comprehensive results for FastGAS and two current baselines, detailing mean, maximum, and minimum values. It is evident that FastGAS consistently delivers superior performance across the majority of trials. Furthermore, FastGAS's performance in the least favorable scenarios is markedly better than that of the baselines in most instances.

\begin{table*}[h]
	\caption{Mean/Maximum/Minimum performance of FastGAS and two baselines across the first four tasks in Table~\ref{tab:main_result} over three trials. The best average performance for each task is highlighted in bold.}
	\centering
	\begin{tabular}{c c c c c c}
		\toprule
            Methods &MRPC &SST-5 &MNLI &DBpedia \\ 
            \midrule
       100 Vote-$k$ &64.60/68.75/62.11  &46.61/47.27/46.09 &38.93/43.75/35.55  &89.19/89.84/88.67  \\
      100 IDEAL &65.49/66.02/64.84  &49.87/52.73/46.09 &41.02/41.41/40.23  &90.63/91.41/89.45  \\
		100 FastGAS &\textbf{66.15}/69.14/62.89  &\textbf{50.26}/55.86/42.58 &\textbf{42.06}/43.75/41.02  &\textbf{90.76}/92.19/88.28  \\ \hline
      18 Vote-$k$ &56.90/67.19/47.27  &41.78/45.70/37.11 &39.45/42.19/37.11  &88.02/91.02/83.59\\
		18 IDEAL &63.80/67.19/59.77  &44.92/48.82/41.79 &39.58/41.80/37.50 &83.20/85.94/81.64  \\
            18 FastGAS &\textbf{65.23}/71.88/55.47  &\textbf{46.61}/48.04/45.70 &\textbf{44.53}/47.66/41.02 &\textbf{88.93}/92.58/84.77   \\
            \midrule
		\bottomrule
	\end{tabular}
	\label{tab:result_first_4}
\end{table*}

\begin{table*}[h]
	\caption{Average/Maximum/Minimum performance of FastGAS and two baselines across the first four tasks in Table~\ref{tab:main_result} over three trials. The best performance for each task is highlighted in bold.}
	\centering
	\begin{tabular}{c c c c c}
		\toprule
            Methods &RTE  &HellaSwag &Xsum   \\ 
            \midrule
       100 Vote-$k$ &57.68/58.20/57.42 &65.89/69.14/64.06 &19.55/19.94/19.13  \\
      100 IDEAL &58.98/60.94/57.42  &64.97/69.53/61.72 &19.68/19.77/19.58   \\
		100 FastGAS &\textbf{61.98}/63.28/60.55  &\textbf{67.45}/71.88/65.23 &\textbf{20.00}/20.55/19.59  \\ \hline
      18 Vote-$k$ &\textbf{56.64}/57.81/55.86  &66.02/71.09/63.28 &19.45/20.45/18.30 \\
		18 IDEAL &53.65/55.47/52.34   &65.89/70.70/63.28  &19.21/20.09/18.76  \\
            18 FastGAS &56.51/57.81/54.69 &\textbf{66.67}/69.92/64.84 &\textbf{20.26}/20.65/19.63\\
            \midrule
		\bottomrule
	\end{tabular}
	\label{tab:result_rest_3}
\end{table*}

\section{Complexity Analysis of FastGAS}\label{sec:complexity}
In this section, we provide the complexity analysis of FastGAS. There are three phases in graph partitioning in FastGAS: the coarsening phase, the partitioning phase, and the uncoarsening phase. In the coarsening phase, we apply a random matching algorithm,  which at most traverses all edges with the complexity $\mathcal{O}(|\mathcal{E}|)$. For the partitioning phase, we apply a breadth-first approach to efficiently bisect the graph, and the complexity of the breadth-first approach $\mathcal{O}(|\mathcal{E}|+|\mathcal{V}|)$. For the uncoarsening phase, the complexity of the KL algorithm can be reduced to $\mathcal{O}(|\mathcal{E}|)$~\cite{fiduccia1988linear}. Thus, the complexity of a 2-way partition is $\mathcal{O}(|\mathcal{E}|+|\mathcal{V}|)$. We bisect the graph log$K$ times to get the $K$-way graph partitioning; the complexity of graph partitioning in FastGAS is $\mathcal{O}((|\mathcal{E}|+|\mathcal{V}|){\rm log}K)$. For the greedy selection in FastGAS, it takes $\mathcal{O}(|\mathcal{V}|/K)$ time to select each node on the subgraphs, thus selecting $M/K$ on each of the $K$ subgraphs needs $\mathcal{O}(|\mathcal{V}|M/K)$, where $M$ is the annotation budgets. In general, the complexity of FastGAS is $\mathcal{O}(|\mathcal{V}|M/K+(|\mathcal{E}|+|\mathcal{V}|){\rm log}K)$, which means the complexity of FastGAS is linear to the size of node set and edge set.

\section{Evaluation on Differeny Text Embedding Models}\label{sec:embedding}
In FastGAS, we use Sentence-BERT as our embedding method and demonstrate the superior performance of FastGAS. In this section, we compare FastGAS with baselines using other text embedding models such as E5~\cite{wang2022text} and InstructOR~\cite{su2022one}. The experiment results of different embedding models are presented below. The findings reveal that FastGAS consistently achieves top performance (five out of six) across different embedding models. This underscores the adaptability of our method, which is effective with various embedding models beyond just Sentence-BERT.

\begin{table*}[h]
	\caption{The average performance of FastGAS and baselines under various text embedding models with the annotation budget of 18.}
	\centering
	\begin{tabular}{c c c c c c}
		\toprule
            Method &Model &MRPC &MNLI &RTE \\ 
            \midrule
       Vote-$k$ &E5 &57.81 &39.97 &55.99   \\
      IDEAL &E5 &\textbf{62.37} &39.58 &55.99  \\
	FastGAS &E5 &61.98 &\textbf{40.49} &\textbf{56.25}  \\ 
            \midrule
      Vote-$k$ &InstructOR &59.77 &40.49 &56.25\\
		IDEAL &InstructOR &63.68 &41.15 &55.73  \\
            FastGAS &InstructOR &\textbf{63.80} &\textbf{41.41} &\textbf{60.29}   \\
            \midrule
		\bottomrule
	\end{tabular}
	\label{tab:embedding}
\end{table*}



\end{document}